\documentclass[12pt]{article}
\usepackage[colorlinks]{hyperref}
\hypersetup{
    colorlinks,
    linkcolor=black,
    citecolor=black,
    filecolor=magenta,
    urlcolor=cyan,
}

\usepackage{resizegather}

\usepackage{cite}
\usepackage{amsmath,amssymb,amsfonts,mathrsfs}
\usepackage{mdframed}
\usepackage{graphicx}
\usepackage{textcomp}
\usepackage[dvipsnames]{xcolor}
\usepackage{tcolorbox}
\usepackage{textcomp}
\usepackage{tablefootnote}
\usepackage{threeparttable}
\usepackage{caption, subcaption}
\usepackage{bbm}
\usepackage{dsfont}
\usepackage{tikz}
\usepackage{graphicx}
\usepackage{tkz-euclide}
\usepackage{algorithm}
\usepackage{algpseudocode}
\usepackage{enumerate}
\usepackage{mathtools}
\usepackage{balance}
\usetikzlibrary{positioning}
\usetikzlibrary{shapes}
\usetikzlibrary{shapes.misc}
\usetikzlibrary{shapes.geometric}
\usetikzlibrary{plotmarks}
\usetikzlibrary{intersections}
\usetikzlibrary{calc}
\usetikzlibrary{fit}
\usetikzlibrary{patterns,tikzmark}
\usetikzlibrary{matrix,decorations.pathreplacing,calc}

\tikzset{cross/.style={cross out, draw, 
         minimum size=2*(#1-\pgflinewidth), 
         inner sep=0pt, outer sep=0pt}}

\newcommand{\vertiii}[1]{{\left\vert\kern-0.25ex\left\vert\kern-0.25ex\left\vert #1 
    \right\vert\kern-0.25ex\right\vert\kern-0.25ex\right\vert}}

\newmdtheoremenv[linecolor=white, backgroundcolor=lightgray!15, innertopmargin=5pt, innerbottommargin=5pt, skipabove=10pt, skipbelow=10pt]{theorem}{\textbf{Theorem}}
\newmdtheoremenv[linecolor=white, backgroundcolor=lightgray!15, innertopmargin=5pt, innerbottommargin=5pt, skipabove=10pt, skipbelow=10pt]{corollary}{\textbf{Corollary}}[theorem]
\newmdtheoremenv[linecolor=white, backgroundcolor=lightgray!15, innertopmargin=5pt, innerbottommargin=5pt, skipabove=10pt, skipbelow=10pt]{lemma}{\textbf{Lemma}}

\newmdtheoremenv[linecolor=white, backgroundcolor=lightgray!15, innertopmargin=5pt, innerbottommargin=5pt, skipabove=10pt, skipbelow=10pt]{problem}{\textbf{Problem}}

\newmdtheoremenv[linecolor=white, backgroundcolor=lightgray!15, innertopmargin=5pt, innerbottommargin=5pt, skipabove=10pt, skipbelow=10pt]{definition}{\textbf{Definition}}

\newmdtheoremenv[linecolor=white, backgroundcolor=lightgray!15, innertopmargin=5pt, innerbottommargin=5pt, skipabove=10pt, skipbelow=10pt]{objective}{\textbf{Objective}}

\newmdenv[
    linecolor=white, backgroundcolor=lightgray!15, innertopmargin=5pt, innerbottommargin=5pt, skipabove=10pt, skipbelow=10pt
]{graybox}

\makeatletter
\renewcommand{\fps@figure}{htp}
\renewcommand{\fps@table}{htp}
\makeatother

\def\BibTeX{{\rm B\kern-.05em{\sc i\kern-.025em b}\kern-.08em
    T\kern-.1667em\lower.7ex\hbox{E}\kern-.125emX}}

\title{RISE: Robust Imitation through Stochastic Encoding}

\author{Mumuksh Tayal, Manan Tayal and Ravi Prakash 
\thanks{All the authors belong to Cyber Physical Systems, Indian Institute of Science (IISc), Bengaluru.
{\{mumukshtayal, manantayal, ravipr\}@iisc.ac.in}.}
}

\begin{document}

\maketitle

\begin{abstract}%
Ensuring safety in robotic systems remains a fundamental challenge, especially when deploying offline policy-learning methods such as imitation learning in dynamic environments. Traditional behavior cloning (BC) often fails to generalize when deployed without fine-tuning as it does not account for disturbances in observations that arises in real-world, changing environments. To address this limitation, we propose RISE (Robust Imitation through Stochastic Encodings), a novel imitation-learning framework that explicitly addresses erroneous measurements of environment parameters into policy learning via a variational latent representation. Our framework encodes parameters such as obstacle state, orientation, and velocity into a smooth variational latent space to improve test time generalization. This enables an offline-trained policy to produce actions that are more robust to perceptual noise and environment uncertainty. We validate our approach on two robotic platforms, an autonomous ground vehicle and a Franka Emika Panda manipulator and demonstrate improved safety robustness while maintaining goal-reaching performance compared to baseline methods.
\end{abstract}

  

\section{Introduction}
\label{section: introduction}

As autonomous robots become increasingly integrated into real-world applications, ensuring safe and high-performance control remains a fundamental challenge. To address safety constraints in such scenarios, various optimal-control strategies have been explored, including Constrained Model Predictive Control (MPC) \cite{MAYNE2000789}, Hamilton–Jacobi (HJ) reachability-based methods \cite{bansal2017hamilton, tayal2025physicsinformedmachinelearningframework}, and Control Barrier Functions (CBFs) \cite{Ames_2017, tayal2024control}. While these approaches provide formal safety guarantees, they typically rely on explicit models of system and environment dynamics, which are often difficult to obtain in real-world settings. Moreover, many of these frameworks assume the ability to perform consistent online rollouts, which raises concerns about the feasibility and safety of conducting unsafe rollouts during training. To address this, several works advocate offline learning approaches \cite{tayal2024semi, leecoptidice} that leverage pre-recorded datasets to avoid repeated unsafe rollouts.

At times, safety concerns also extend to demonstration data, where recording unsafe demonstrations may be infeasible. In such cases, Imitation Learning (IL) is a promising approach for training control policies from safe expert demonstrations, particularly when system dynamics are partially known or difficult to model. However, traditional IL methods such as behavioral cloning (BC) often struggle to generalize beyond the training distribution, resulting in degraded performance at deployment, especially when measurement equipment provides only rough estimates of obstacle positions and the surrounding environment. Several techniques propose adding noise during demonstration collection \cite{Laskey2017DART, ciftci2024safe} to encourage robustness, but this is not always possible, particularly when the demonstration data is pre-recorded.

Attempts have been made to combine methods such as Constrained MPC and CBFs with IL \cite{Robey2020OfflineNCBF, tayal2025cp} to improve constraint satisfaction in robotic systems. For example, HJ reachability–based imitation learning \cite{ciftci2024safe} enforces control constraints by computing forward or backward reachable sets to guarantee safety. However, these methods are computationally expensive and do not scale well to high-dimensional robotic systems due to the curse of dimensionality. Similarly, some CBF-based IL approaches \cite{cosner2022end} do not explicitly enforce control bounds and often assume unlimited control authority, which is impractical for real systems.

In many practical applications, key safety-related cues (e.g., obstacle position, velocity, and geometry) are available from onboard  (or even precise coordinates through motion capture), and task specifications such as goal positions are often pre-defined and may vary across deployments. Although these structured environmental cues are available, their precise and accurate state coordinates and velocity may not be obtainable in real time; instead, only rough measured estimates are typically available.

\begin{figure}[t]
\centering
\includegraphics[width=0.9\linewidth]{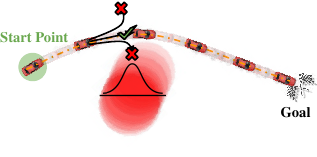}
\caption{Behavior Cloning (BC) struggles to generalize for noisy obstacle readings in real-world and often collides with dynamic obstacles, whereas \textbf{RISE} picks safer actions without deviating from the original trajectory while it  navigates through dynamic obstacles.}
\label{fig: intro-image}
\vspace{-1em}
\end{figure}

To address these challenges, we propose RISE, a novel framework that accounts for noisy measurements of obstacle data by conditioning the policy on a learned probabilistic latent space. Specifically, we build on Goal-Conditioned Imitation Learning (GCIL) \cite{ding2020gcil, reuss2023goal}, and augment it by integrating the variational encoder to accommodate noisy, yet safety-critical environmental factors into a structured latent representation. This enables a more realistic interpolation between environment parameters, thus, improving task adaptability for unseen intermediate datapoints while leveraging the structured perturbation data directly from the latent space. This improves inculcating inherent awareness of safety from the provided data, which is generally hard to achieve with behavior cloning alone.

To summarize, the key contributions of our paper are:
\begin{itemize}
\item Unlike CBF and HJ reachability approaches, which require exact system dynamics or an accurate environment model, RISE operates in real-world settings where dynamics are unavailable or imprecise.
\item We train a variational autoencoder (VAE) to predict a probabilistic distribution over obstacle states, thereby capturing uncertainty from noisy observations. Conditioning the policy on this distribution yields more robust behaviors that avoid likely obstacle locations.
\item Few safety-critical policy-learning frameworks (including many CBF-based methods) explicitly consider physical actuation limits. Our approach incorporates actuation constraints and learns policies that respect those limits while behaving conservatively near obstacles.
\item We validate the framework in simulated robotic environments and on hardware. Comparative analyses against baselines such as PCIL and C-PPO \cite{10.5555/3305381.3305384, alshiekh2018safe} show improved safety with maintained goal-reaching performance.
\end{itemize}


\section{Preliminaries} 
\label{section: preliminaries}
\subsection{Safe Imitation Learning}
Most regular Imitation learning frameworks train policies that map observations to actions using expert demonstrations, in cases where dynamics of the system is unknown or very complex. Various IL approaches, including behavioral cloning (BC) \cite{pomerleau1991efficient}, DAgger \cite{ross2011reductionimitationlearningstructured}, and inverse reinforcement learning (IRL) \cite{ng2000algorithms} exist, however, even though they are derived from the demonstrations of a superior policy, they lack the capability to learn safety-aware policies due to lack of true reward signals. Safe Imitation Learning frameworks, on the other hand, either try to remain within in-distribution region as demonstrated by the expert demonstrations \cite{castaneda2023idbf, Robey2020OfflineNCBF}, thus, minimizing the risk of violating safety, or they add an adversarial noise to the demonstrations while recording them \cite{ciftci2024safe, Laskey2017DART} to inherently learn a more robust policy. But it is important to also consider that these approaches have their shortcomings, either they are overly conservative or they require a specific dataset to train, both of which may not be acceptable at all times.

\subsection{Parameter-Conditioned Imitation Learning}
It is a subdomain of Imitation Learning, where each demonstration data-point is augmented with one or more parameters (e.g., goal state \cite{ding2020gcil}), hence, seeking to obtain the indicator reward for the task that the demonstration was provided for. The conditioning parameter contains information that a learning method can leverage to disambiguate demonstrations. Parameters such as goal-states have also extended the domain of reinforcement learning through Goal Conditioned Reinforcement Learning (GCRL) \cite{schaul2015uvfa}, where the agent is not provided expert demonstrations but reward signals instead. Typically these reward signals are difficult to define, especially for complex tasks and environments, providing demonstrations is often a more natural option in such situations. Additionally, the policy rollouts required by GCRL are often expensive in real-world settings.

\subsection{Variational AutoEncoder (VAE)}
Variational AutoEncoders (VAEs) \cite{kingma2013auto} are generative models that learn a probabilistic latent space representation of data. A VAE consists of an encoder and a decoder component, both of which are connected to each other using the reparameterization trick (as referred in \cite{kingma2013auto}).


The objective function of a VAE is to maximize the Evidence Lower Bound (ELBO):
\begin{equation}
    \mathcal{ELBO}(x) = \mathbb{E}_{q_\phi(z|x)}[\log p_\theta(x|z)] - D_{KL}(q_\phi(z|x) || p(z)),
\end{equation}
where the first term maximizes the likelihood of reconstructions, and the second term regularizes the latent space by minimizing the Kullback-Leibler (KL) divergence between the approximate posterior and a prior distribution \( p(z) \).

VAEs have been widely adopted for learning structured representations, denoising, and improving generalization in downstream tasks, making them a valuable tool for enhancing imitation learning in RL \cite{higgins2017beta}, thus, making them an ideal choice for an application like ours.

\section{Methodology}
\label{section: method}
\begin{figure*}[ht]
\centering
    \includegraphics[width=0.99\textwidth]{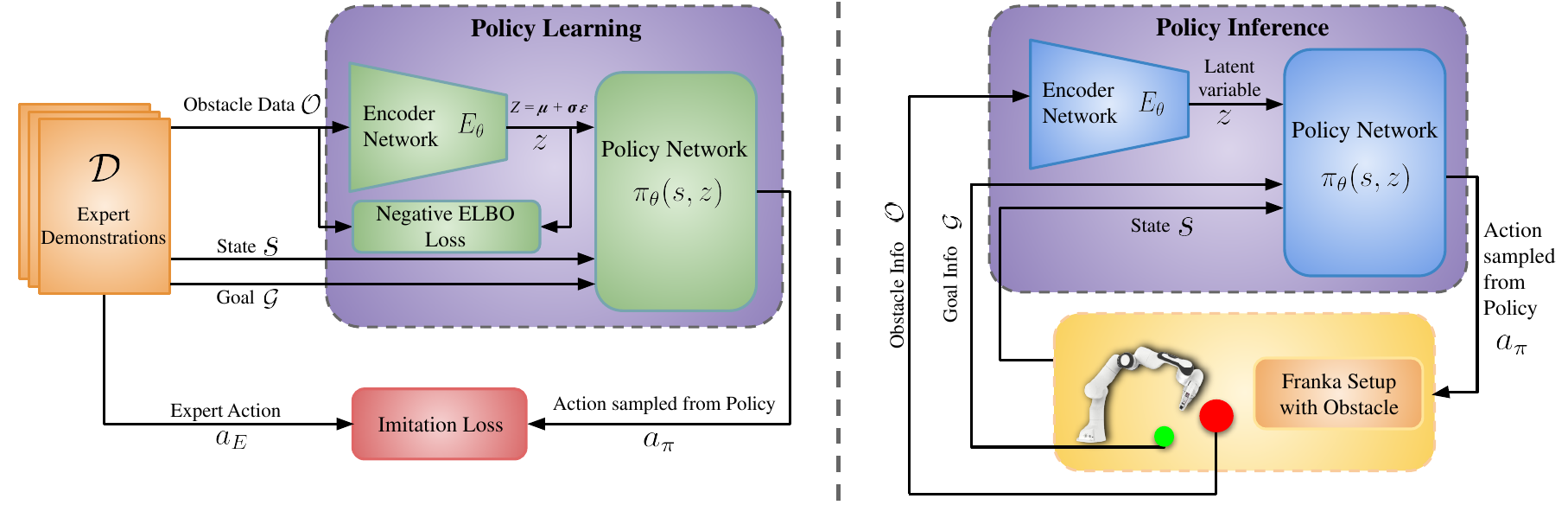}
    \caption{\textbf{Architecture:} \textit{(Left)} During training, expert demonstration trajectories, augmented with obstacle parameters, and goal coordinates, are used to learn the Variational Encoder to produce a latent representation (z), which is combined with agent’s state to drive training of the Policy Network. \textit{(Right)} At inference, trained architecture deploys the Policy Network on a real agent, generating actions for safe navigation in dynamic environments.}
    \label{fig:framework}
    \vspace{-10pt}
\end{figure*}

In this section, we present the framework that learns a latent unsafe region distribution for given noisy obstacle perception signals to enable robust imitation learning. The method first encodes raw measured parameters (e.g., obstacle position, obstacle velocity, and obstacle radius as in the cases demonstrated) into a structured latent variable, and then conditions an imitation policy on the current state, goal region and this derived latent distribution.

\subsection{Problem Formulation}
Consider a robotic system with state space $\mathcal{S}$, action space $\mathcal{A}$, and a set of safety parameters $\mathcal{C}$. The safety parameters $c \in \mathcal{C}$ represent critical environmental features such as obstacle positions, velocities, and geometries. The objective is to learn a policy $\pi : \mathcal{S} \times \mathcal{C} \rightarrow \mathcal{A}$ that maps states and safety parameters to actions while maintaining safety constraints and accomplishing the desired task.

\subsection{Latent Unsafe Region}
Let $c \in \mathbb{R}^{d_c}$ denote the raw measured parameters. We employ a variational encoder network $E(\cdot)$ to embed $c$ into a latent normal distribution. To train our VAE-style encoder using ELBO, we model $z$ using the reparameterization trick:
\begin{equation}
    z = \mu + \sigma \odot \epsilon, \quad \epsilon \sim \mathcal{N}(0, I),
\end{equation}
where $\mu$ is the mean and $\sigma$ is the deviation, which are the outputs of the encoder network. Thence derived latent variable $z$ is passed into the decoder and the VAE is then learnt using the negative ELBO loss.

The encoder architecture consists of fully connected layers with ReLU activations, culminating in parallel output layers for $\mu$ and $\sigma$.

\subsection{Behavior Policy}
The policy $\pi(s,g,z)$ maps the current state $s \in \mathbb{R}^{d_s}$, the goal $g$ and the latent safety variable $z$ to an action $a \in \mathbb{R}^{d_a}$:
\begin{equation}
    a = \pi(s, g, z).
\end{equation}
Unlike traditional behavior cloning approaches that directly map states to actions, our policy also leverages the structured latent representation of measured obstacle parameters.
By sampling from the VAE posterior during training, the policy effectively sees multiple plausible obstacle hypotheses (a form of virtual data augmentation), which improves robustness to perceptual variation.
The policy network uses a 2-layered fully connected Neural Network with 128 neurons in the hidden layer with ReLU activations. The state vector $s$, goal vector $g$ and latent vector $z$ are concatenated and passed through these layers to produce the action output.

Algorithm~\ref{alg:train} summarizes the training procedure where it integrates the encoder-decoder architecture with the policy network in an end-to-end training framework.

\begin{algorithm}[ht!]
\caption{Training Algorithm}
\begin{algorithmic}[1]
\Require Dataset $\mathcal{D}=\{(s,a,s',\mathrm{obs})\}$
\State \textbf{Stage 1: Pre-Train VAE (ELBO)}
\For{epoch = 1 \textbf{to} $E_{\mathrm{VAE}}$}
  \For{batch $\{\mathrm{obs}\}$ from $\mathcal{D}$}
    \State $\mu,\sigma \leftarrow q_\phi(\mathrm{obs})$
    \State $\epsilon\sim\mathcal{N}(0,I),\; z=\mu+\sigma\odot\epsilon$
    \State $\hat{\mathrm{obs}}\leftarrow p_\psi(z)$
    \State $\mathcal{L} \leftarrow -\log p_\psi(\hat{\mathrm{obs}}|\mathrm{z}) + \beta\,\mathrm{KL}[q_\phi(z|\mathrm{obs})\|p(z)]$
    \State Update $\phi,\psi \leftarrow \nabla_{\phi,\psi}\mathcal{L}_{\text{ELBO}}$
  \EndFor
\EndFor
\State Freeze VAE parameters: $\phi,\psi \leftarrow \text{detach}$
\Statex
\State \textbf{Stage 2: Train NN policy}
\For{epoch = 1 \textbf{to} $E_{\pi}$}
  \For{batch $\{(s,a,g,\mathrm{obs})\}$ from $\mathcal{D}$}
    \State $\mu,\sigma \leftarrow q_\phi(\mathrm{obs})$ \Comment{no gradients into VAE}
    \State \(\displaystyle \mathcal{L}_{\pi}\leftarrow 0\)
    \For{$m=1\ldots M$} 
    \Comment{M unique perturbations}
      \State $\epsilon^{(m)}\sim\mathcal{N}(0,I),\; z^{(m)}=\mu+\sigma\odot\epsilon^{(m)}$
      \State $\hat a^{(m)}\leftarrow \pi_\theta(s,g,z^{(m)})$
      \State $\mathcal{L}_{\pi} \mathrel{+}= \ell\big(\hat a^{(m)},a\big)$ \Comment{e.g., MSE / NLL}
    \EndFor
    \State $\mathcal{L}_{\pi}\leftarrow \mathcal{L}_{\pi}/M$
    \State Update $\theta \leftarrow \nabla_{\theta}\mathcal{L}_{\pi}$
  \EndFor
\EndFor
\State \Return trained policy $\pi_\theta$ (VAE used in inference to sample $z$)
\end{algorithmic}
\label{alg:train}
\end{algorithm}


\section{Experiments}
\label{section: experiments}
In our experiments we ask whether the proposed method can reliably handle inputs that lie within the data distribution yet were not observed during training, i.e., whether the policy can interpolate across realistic, unseen environment configurations produced by noisy measurements. In particular, we evaluate how well the framework mitigates distribution shift arising from disturbances in sensor or tracking readings. We benchmark against representative baselines on two simulated tasks (autonomous navigation of a ground vehicle and a Reach-Safe Franka Emika Panda manipulation task) and demonstrate results on Franka Panda hardware. Our evaluation emphasizes safety metrics and robustness under randomized initial conditions, and in the following subsections we describe how we generate challenging test cases for thorough performance assessment.
\begin{figure}[]
    \centering
    \includegraphics[width=0.99\linewidth]{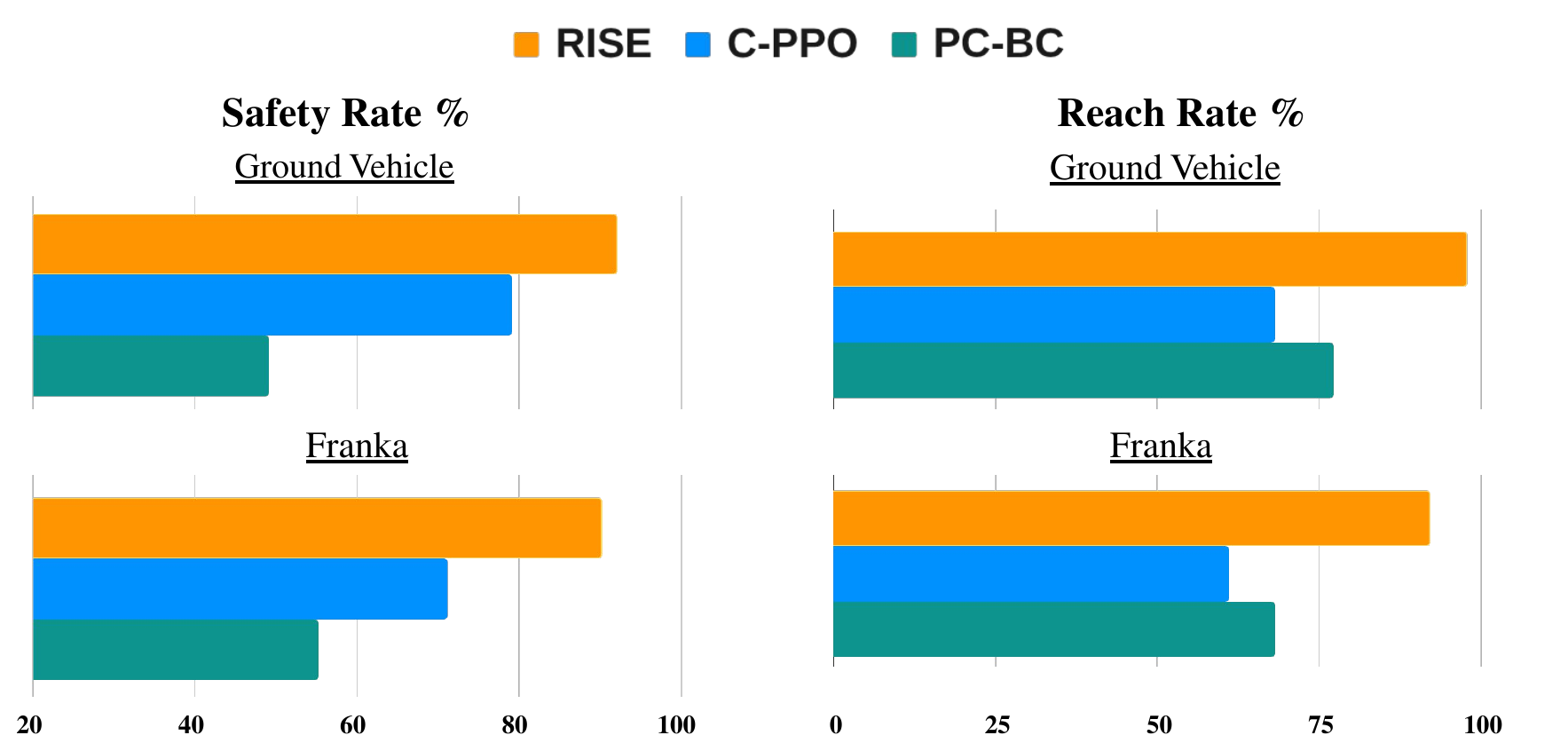}
\caption{Comparative study between all baselines and our approach based on evaluation matrics. The top plots illustrate the results for the Ground Vehicle Navigation setup, meanwhile, the bottom plots correspond to Franka Manipulator Task results from simulation.
\vspace{-1.5em}
}
\label{fig: Baseline_comparison}
\end{figure}

\subsection{Baselines and Evaluation Metrics}
To evaluate our framework, we compare it against two baselines: Parameter-Conditioned Behavior Cloning (PC-BC) \cite{ding2020gcil}, which learns policies through behavior cloning with explicit conditioning on environmental parameters under randomized safety conditions, relying only on domain randomization for generalization. Primarily, we use the same inputs as our proposed approach, just without the Variational Encoder in this case; and Constrained Proximal Policy Optimization (C-PPO) \cite{pmlr-v119-stooke20a}, which extends PPO by incorporating safety constraints using Lagrange multipliers to penalize constraint violations during training. PC-BC tests whether including the variational encoder affects the performance or just a raw exposure to diverse conditions alone enables generalization, while C-PPO provides a reinforcement learning-based comparison that explicitly incorporates constraints. The performance of our method and its baselines are assessed using the following two key metrics:
\begin{enumerate}
    \item \textbf{Safety Rate}: Percentage of test trials in which the agent doesn't collide with the obstacle at any timestamp.
    \item \textbf{Reach Rate}: Percentage of complete trials where the learned policy successfully reaches the goal. Note that a successfully reached episode is one during which the agent doesn't collide into the obstacle at any point in time during the entire episode.
\end{enumerate}

\subsection{Autonomous Navigation of a Ground Vehicle}


In this task, the agent, an autonomous ground vehicle must reach the goal while navigating an environment containing a dynamic obstacle. The agent's state is represented as $s = (x, y, \theta)$, where $(x, y)$ denotes position and $\theta$ is orientation. The action space consists of linear and angular velocities, $a = (v, \omega)$.
The environment features a moving obstacle whose position is sampled to ensure a safe margin between the agent's initial state and goal. The obstacle’s radius varies in a range of values to introduce variability, and its velocity is dynamically assigned to create unpredictable motion.

\subsection{Franka Manipulator Task}
In this task, a Franka Panda manipulator must reach a parameterized goal while avoiding obstacles in its workspace. For this experiment, we have used safe-panda-gym simulation environment \cite{SafePandaGym, gallouedec2021pandagym}. 
The action space comprises end-effector displacement, $a = (dx, dy, dz)$, applied through Position Control.

\subsection{Training Data \& Evaluation}
\textbf{Expert Data Generation:} Expert demonstrations are generated using a mixture of experts like model predictive control \cite{MAYNE2000789}, control barrier function (CBF), etc. The dataset, which includes 10k expert demonstrations, is constructed by randomly sampling initial robot states, and goal position. 

\textbf{Evaluation and Comparative Analysis:}  
We evaluate performance across 1000 sampled test scenarios, all sampled by adding random noise (sampled from standard normal distribution) to the training data to emulate the desired noise in environmental parameter readings. Figure~\ref{fig: Baseline_comparison} summarizes the results. Our approach outperforms both C-PPO and PC-BC on both the evaluation metrics. It achieves the highest \textbf{Safety Rate} while maintaining a superior \textbf{Reach Rate}. Although C-PPO comes close in terms of safety, it struggles with goal-reaching performance, thus showing its conservative nature. On the other hand, we see PC-BC to be more aggressive and hence, suffers from frequent collisions. These results underscore the ability of RISE to balance safety while maintaining task performance.

\begin{figure}[]
    \centering
    \includegraphics[width=0.99\linewidth]{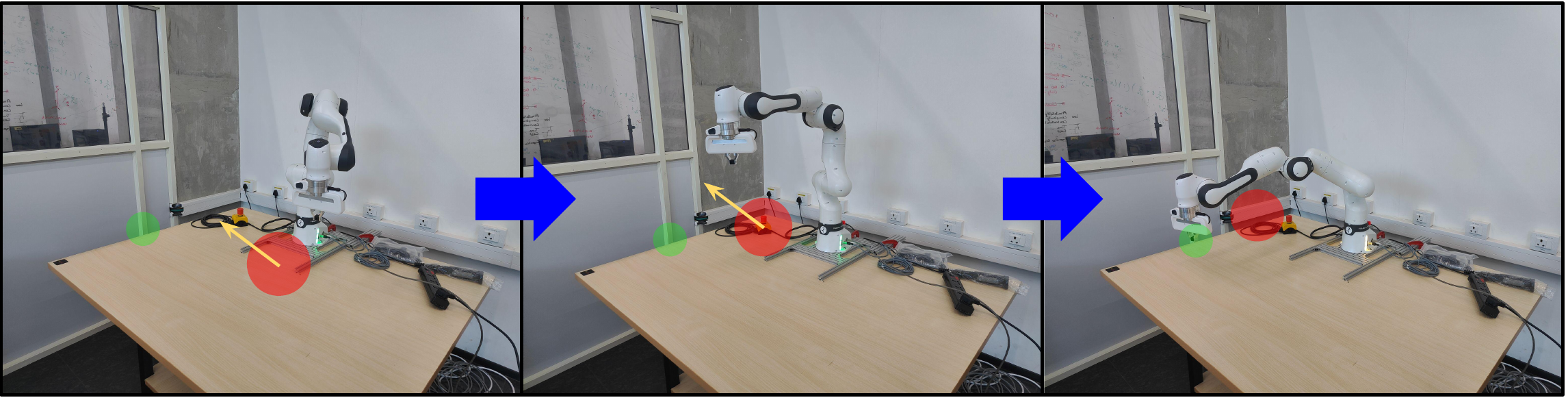}
    \caption{Illustration shows \textbf{Franka Panda manipulator} advancing toward its designated target (\textcolor{ForestGreen}{green region}) while executing collision avoidance maneuvers in the presence of a dynamic obstacle (\textcolor{red}{red sphere}). \textcolor{YellowOrange}{Arrows} indicate direction.}
    \label{fig:Franka_key_frame}
\end{figure}



\textbf{Hardware Results:} We further validate our framework on a physical Franka Emika Panda manipulator. In this setup, virtual obstacles are employed, and environmental parameters (obstacle properties and goal locations) are provided in real time to the policy. Figure \ref{fig:Franka_key_frame} presents key frames from the hardware demonstration, confirming successful goal-reaching with effective obstacle avoidance.

\section{Conclusion}
\label{section: conclusions}
In this paper, we proposed a practical imitation-learning framework that makes policies robust to realistic measurement uncertainty by conditioning them on a variational latent representation of environment parameters. By sampling plausible obstacle states from the VAE posterior during training, the policy learns to interpolate across nearby, realistic percepts and therefore behaves more conservatively and reliably in noisy, dynamic scenes without requiring explicitly requiring to train the model on these datapoints, nor requiring the exact dynamics models. Validation on an autonomous ground vehicle and a Franka Emika Panda demonstrates improved safety while preserving goal-reaching performance versus baselines. 
\newpage

\bibliographystyle{IEEEtran}
\bibliography{ref.bib}

\begin{thebibliography}{10}
\providecommand{\url}[1]{#1}
\csname url@samestyle\endcsname
\providecommand{\newblock}{\relax}
\providecommand{\bibinfo}[2]{#2}
\providecommand{\BIBentrySTDinterwordspacing}{\spaceskip=0pt\relax}
\providecommand{\BIBentryALTinterwordstretchfactor}{4}
\providecommand{\BIBentryALTinterwordspacing}{\spaceskip=\fontdimen2\font plus
\BIBentryALTinterwordstretchfactor\fontdimen3\font minus \fontdimen4\font\relax}
\providecommand{\BIBforeignlanguage}[2]{{%
\expandafter\ifx\csname l@#1\endcsname\relax
\typeout{** WARNING: IEEEtran.bst: No hyphenation pattern has been}%
\typeout{** loaded for the language `#1'. Using the pattern for}%
\typeout{** the default language instead.}%
\else
\language=\csname l@#1\endcsname
\fi
#2}}
\providecommand{\BIBdecl}{\relax}
\BIBdecl

\bibitem{MAYNE2000789}
\BIBentryALTinterwordspacing
D.~Mayne, J.~Rawlings, C.~Rao, and P.~Scokaert, ``Constrained model predictive control: Stability and optimality,'' \emph{Automatica}, vol.~36, no.~6, pp. 789--814, 2000. [Online]. Available: \url{https://www.sciencedirect.com/science/article/pii/S0005109899002149}
\BIBentrySTDinterwordspacing

\bibitem{bansal2017hamilton}
S.~Bansal, M.~Chen, S.~Herbert, and C.~J. Tomlin, ``Hamilton-jacobi reachability: A brief overview and recent advances,'' in \emph{2017 IEEE 56th Annual Conference on Decision and Control (CDC)}.\hskip 1em plus 0.5em minus 0.4em\relax IEEE, 2017, pp. 2242--2253.

\bibitem{tayal2025physicsinformedmachinelearningframework}
\BIBentryALTinterwordspacing
M.~Tayal, A.~Singh, S.~Kolathaya, and S.~Bansal, ``A physics-informed machine learning framework for safe and optimal control of autonomous systems,'' 2025. [Online]. Available: \url{https://arxiv.org/abs/2502.11057}
\BIBentrySTDinterwordspacing

\bibitem{Ames_2017}
A.~D. Ames, X.~Xu, J.~W. Grizzle, and P.~Tabuada, ``Control barrier function based quadratic programs for safety critical systems,'' \emph{{IEEE} Transactions on Automatic Control}, vol.~62, no.~8, pp. 3861--3876, 2017.

\bibitem{tayal2024control}
M.~Tayal, R.~Singh, J.~Keshavan, and S.~Kolathaya, ``Control barrier functions in dynamic uavs for kinematic obstacle avoidance: A collision cone approach,'' in \emph{2024 American Control Conference (ACC)}.\hskip 1em plus 0.5em minus 0.4em\relax IEEE, 2024, pp. 3722--3727.

\bibitem{tayal2024semi}
M.~Tayal, A.~Singh, P.~Jagtap, and S.~Kolathaya, ``Semi-supervised safe visuomotor policy synthesis using barrier certificates,'' \emph{arXiv preprint arXiv:2409.12616}, 2024.

\bibitem{leecoptidice}
J.~Lee, C.~Paduraru, D.~J. Mankowitz, N.~Heess, D.~Precup, K.-E. Kim, and A.~Guez, ``Coptidice: Offline constrained reinforcement learning via stationary distribution correction estimation,'' in \emph{International Conference on Learning Representations}, 2022.

\bibitem{Laskey2017DART}
\BIBentryALTinterwordspacing
M.~Laskey, J.~Lee, R.~Fox, A.~Dragan, and K.~Goldberg, ``Dart: Noise injection for robust imitation learning,'' in \emph{Proceedings of the 1st Annual Conference on Robot Learning}, ser. Proceedings of Machine Learning Research, S.~Levine, V.~Vanhoucke, and K.~Goldberg, Eds., vol.~78.\hskip 1em plus 0.5em minus 0.4em\relax PMLR, 13--15 Nov 2017, pp. 143--156. [Online]. Available: \url{https://proceedings.mlr.press/v78/laskey17a.html}
\BIBentrySTDinterwordspacing

\bibitem{ciftci2024safe}
Y.~U. Ciftci, D.~Chiu, Z.~Feng, G.~S. Sukhatme, and S.~Bansal, ``Safe-gil: Safety guided imitation learning for robotic systems,'' \emph{arXiv preprint arXiv:2404.05249}, 2024.

\bibitem{Robey2020OfflineNCBF}
A.~Robey, H.~Hu, L.~Lindemann, H.~Zhang, D.~V. Dimarogonas, S.~Tu, and N.~Matni, ``Learning control barrier functions from expert demonstrations,'' in \emph{2020 59th IEEE Conference on Decision and Control (CDC)}, 2020, pp. 3717--3724.

\bibitem{tayal2025cp}
M.~Tayal, A.~Singh, P.~Jagtap, and S.~Kolathaya, ``Cp-ncbf: A conformal prediction-based approach to synthesize verified neural control barrier functions,'' \emph{arXiv preprint arXiv:2503.17395}, 2025.

\bibitem{cosner2022end}
R.~K. Cosner, Y.~Yue, and A.~D. Ames, ``End-to-end imitation learning with safety guarantees using control barrier functions,'' in \emph{2022 IEEE 61st Conference on Decision and Control (CDC)}.\hskip 1em plus 0.5em minus 0.4em\relax IEEE, 2022, pp. 5316--5322.

\bibitem{ding2020gcil}
\BIBentryALTinterwordspacing
Y.~Ding, C.~Florensa, P.~Abbeel, and M.~Phielipp, ``Goal-conditioned imitation learning,'' in \emph{Advances in Neural Information Processing Systems}, H.~Wallach, H.~Larochelle, A.~Beygelzimer, F.~d\textquotesingle Alch\'{e}-Buc, E.~Fox, and R.~Garnett, Eds., vol.~32.\hskip 1em plus 0.5em minus 0.4em\relax Curran Associates, Inc., 2019. [Online]. Available: \url{https://proceedings.neurips.cc/paper_files/paper/2019/file/c8d3a760ebab631565f8509d84b3b3f1-Paper.pdf}
\BIBentrySTDinterwordspacing

\bibitem{reuss2023goal}
M.~Reuss, M.~Li, X.~Jia, and R.~Lioutikov, ``Goal-conditioned imitation learning using score-based diffusion policies,'' \emph{arXiv preprint arXiv:2304.02532}, 2023.

\bibitem{10.5555/3305381.3305384}
J.~Achiam, D.~Held, A.~Tamar, and P.~Abbeel, ``Constrained policy optimization,'' in \emph{Proceedings of the 34th International Conference on Machine Learning - Volume 70}, ser. ICML'17.\hskip 1em plus 0.5em minus 0.4em\relax JMLR.org, 2017, p. 22–31.

\bibitem{alshiekh2018safe}
\BIBentryALTinterwordspacing
M.~Alshiekh, R.~Bloem, R.~Ehlers, B.~Könighofer, S.~Niekum, and U.~Topcu, ``Safe reinforcement learning via shielding,'' \emph{Proceedings of the AAAI Conference on Artificial Intelligence}, vol.~32, no.~1, Apr. 2018. [Online]. Available: \url{https://ojs.aaai.org/index.php/AAAI/article/view/11797}
\BIBentrySTDinterwordspacing

\bibitem{pomerleau1991efficient}
D.~A. Pomerleau, ``Efficient training of artificial neural networks for autonomous navigation,'' \emph{Neural computation}, vol.~3, no.~1, pp. 88--97, 1991.

\bibitem{ross2011reductionimitationlearningstructured}
\BIBentryALTinterwordspacing
S.~Ross, G.~J. Gordon, and J.~A. Bagnell, ``A reduction of imitation learning and structured prediction to no-regret online learning,'' 2011. [Online]. Available: \url{https://arxiv.org/abs/1011.0686}
\BIBentrySTDinterwordspacing

\bibitem{ng2000algorithms}
A.~Y. Ng and S.~J. Russell, ``Algorithms for inverse reinforcement learning,'' in \emph{Proceedings of the Seventeenth International Conference on Machine Learning}.\hskip 1em plus 0.5em minus 0.4em\relax San Francisco, CA, USA: Morgan Kaufmann Publishers Inc., 2000, p. 663–670.

\bibitem{castaneda2023idbf}
\BIBentryALTinterwordspacing
F.~Casta\~neda, H.~Nishimura, R.~T. McAllister, K.~Sreenath, and A.~Gaidon, ``In-distribution barrier functions: Self-supervised policy filters that avoid out-of-distribution states,'' in \emph{Proceedings of The 5th Annual Learning for Dynamics and Control Conference}, ser. Proceedings of Machine Learning Research, N.~Matni, M.~Morari, and G.~J. Pappas, Eds., vol. 211.\hskip 1em plus 0.5em minus 0.4em\relax PMLR, 15--16 Jun 2023, pp. 286--299. [Online]. Available: \url{https://proceedings.mlr.press/v211/castaneda23a.html}
\BIBentrySTDinterwordspacing

\bibitem{schaul2015uvfa}
\BIBentryALTinterwordspacing
T.~Schaul, D.~Horgan, K.~Gregor, and D.~Silver, ``Universal value function approximators,'' in \emph{Proceedings of the 32nd International Conference on Machine Learning}, ser. Proceedings of Machine Learning Research, F.~Bach and D.~Blei, Eds., vol.~37.\hskip 1em plus 0.5em minus 0.4em\relax Lille, France: PMLR, 07--09 Jul 2015, pp. 1312--1320. [Online]. Available: \url{https://proceedings.mlr.press/v37/schaul15.html}
\BIBentrySTDinterwordspacing

\bibitem{kingma2013auto}
\BIBentryALTinterwordspacing
D.~P. Kingma and M.~Welling, ``Auto-encoding variational bayes,'' 2022. [Online]. Available: \url{https://arxiv.org/abs/1312.6114}
\BIBentrySTDinterwordspacing

\bibitem{higgins2017beta}
I.~Higgins, L.~Matthey, A.~Pal, C.~Burgess, X.~Glorot, M.~Botvinick, S.~Mohamed, and A.~Lerchner, ``beta-vae: Learning basic visual concepts with a constrained variational framework,'' in \emph{International conference on learning representations}, 2017.

\bibitem{pmlr-v119-stooke20a}
\BIBentryALTinterwordspacing
A.~Stooke, J.~Achiam, and P.~Abbeel, ``Responsive safety in reinforcement learning by {PID} lagrangian methods,'' in \emph{Proceedings of the 37th International Conference on Machine Learning}, ser. Proceedings of Machine Learning Research, H.~D. III and A.~Singh, Eds., vol. 119.\hskip 1em plus 0.5em minus 0.4em\relax PMLR, 13--18 Jul 2020, pp. 9133--9143. [Online]. Available: \url{https://proceedings.mlr.press/v119/stooke20a.html}
\BIBentrySTDinterwordspacing

\bibitem{SafePandaGym}
S.~W. Tosin~Oseni, ``Safe panda gym,'' \url{https://github.com/tohsin/Safe-panda-gym}, 2022.

\bibitem{gallouedec2021pandagym}
Q.~Gallou{\'e}dec, N.~Cazin, E.~Dellandr{\'e}a, and L.~Chen, ``{panda-gym: Open-Source Goal-Conditioned Environments for Robotic Learning},'' \emph{4th Robot Learning Workshop: Self-Supervised and Lifelong Learning at NeurIPS}, 2021.

\end{thebibliography}

\end{document}